\pdfoutput=1

\documentclass[11pt]{article}

\usepackage{acl}

\usepackage{times}
\usepackage{latexsym}

\usepackage[T1]{fontenc}

\usepackage[utf8]{inputenc}

\usepackage{microtype}

\usepackage{graphicx}
\usepackage{todonotes}
\usepackage{mathtools, nccmath}
\usepackage{url}
\usepackage{amsfonts}
\usepackage{array}
\usepackage{subcaption}
\usepackage{booktabs}

\usepackage{cleveref}

\crefformat{section}{\S#2#1#3}
\crefformat{subsection}{\S#2#1#3}
\crefformat{subsubsection}{\S#2#1#3}
\crefrangeformat{section}{\S\S#3#1#4 to~#5#2#6}
\crefmultiformat{section}{\S\S#2#1#3}{ and~#2#1#3}{, #2#1#3}{ and~#2#1#3}

\usepackage{microtype}

\newcommand\given[1][]{\:#1\vert\:}

\title{CORWA: A Citation-Oriented Related Work Annotation Dataset}

\author{Xiangci Li ~~ Biswadip Mandal ~~ Jessica Ouyang\\
  Department of Computer Science \\
  University of Texas at Dallas \\
  Richardson, TX 75080 \\
  \tt lixiangci8@gmail.com, \\ 
  \tt \{Biswadip.Mandal, Jessica.Ouyang\}@UTDallas.edu
}

\begin{document}
\maketitle
\begin{abstract}
Academic research is an exploratory activity to discover new solutions to problems. By this nature, academic research works perform literature reviews to distinguish their novelties from prior work. In natural language processing, this literature review is usually conducted under the “Related Work” section. The task of related work generation aims to automatically generate the related work section given the rest of the research paper and a list of papers to cite. Prior work on this task has focused on the sentence as the basic unit of generation, neglecting the fact that related work sections consist of variable length text fragments derived from different information sources. As a first step toward a linguistically-motivated related work generation framework, we present a Citation Oriented Related Work Annotation (CORWA) dataset that labels different types of citation text fragments from different information sources. We train a strong baseline model that automatically tags the CORWA labels on massive unlabeled related work section texts. We further suggest a novel framework for human-in-the-loop, iterative, abstractive related work generation. 
\end{abstract}

\section{Introduction}

Academic research is an exploratory activity to solve problems that have never been solved before. By this nature, each academic research work must sit at the frontier of its field and present novel contributions that have not been addressed by prior work; in order to convince readers of the novelty of the current work, the authors must compare against the prior work. While the format may vary among different fields, in natural language processing (NLP), this literature review is usually conducted under the “Related Work” section. Since each paper must review the relevant prior work in its field, which is shared among papers on the same topic or task, many related work sections in a given field can be similar in both content and format. Therefore, it is a natural motivation to develop a system for generating related work sections automatically. 

\begin{figure}
\centering
    \includegraphics[width=0.45\textwidth]{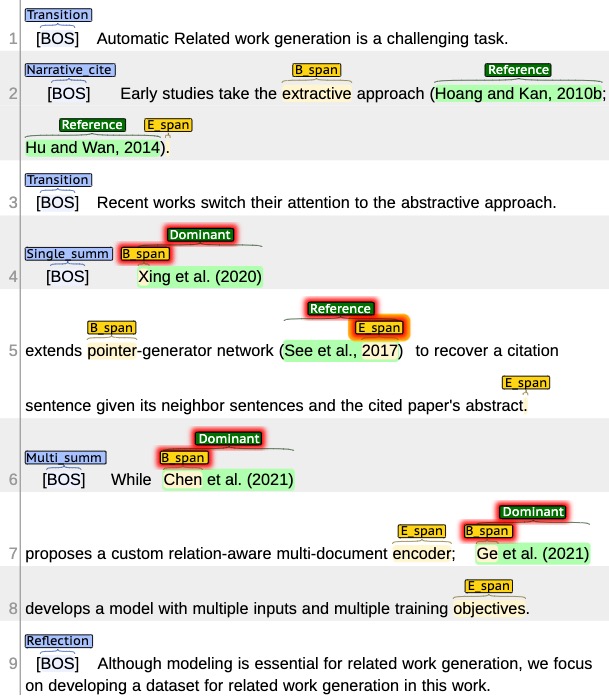}
     \vspace{-0.5em}
  \caption{An example of CORWA labels displayed using the BRAT interface \cite{stenetorp2012brat}.} 
  \label{fig:CORWA_example}
  \vspace{-1em}
\end{figure}

The task of automatic related work generation is that of generating the related work section of a target paper given the rest of the target paper and a set of papers to cite. 
Prior works \cite{hoang2010towards, hu2014automatic, chen2019automatic, wang2019toc, xing2020automatic, ge-etal-2021-baco, luu-etal-2021-explaining, chen-etal-2021-capturing} mostly simplify related work generation as a general summarization task, generating related work sections using sentence-level models. This approach ignores the nature of the related work section, which consists of variable-length text fragments derived from different information sources. These text fragments refer to different cited papers, and they range in length from a few words to multiple sentences. There are also non-citation, supporting sentences that serve various discursive roles, such as introducing new topics, transitioning between topics, or reflecting on the current work. We argue it is necessary to distinguish these heterogeneous text fragments, rather than treating related work sections as concatenations of homogeneous sentences.

In addition to the heterogeneous information sources for related work section sentences, the writing styles of these sentences also vary. \citet{khoo2011analysis} classify literature reviews to be integrative or descriptive, depending on whether they focus on high-level ideas or provide more detailed information on specific studies. However, this document-level classification scheme was intended as a descriptive, information science study of related work sections, and it has not been previously used in automatic related work generation.

Inspired by these observations, as a first step towards linguistically-motivated related work generation, we present a Citation Oriented Related Work Annotation (CORWA) dataset of related work sections from NLP papers. We distinguish text fragments from different information sources by tagging each sentence with discourse labels 
and identifying the spans of tokens belonging to each citation. We further distinguish citations that give detailed explanations of cited papers and those that illustrate high-level concepts. 

Our main contributions are as follows: (1) We collect a CORWA dataset that decomposes the related work section with three inter-related annotation tasks --- discourse tagging, citation span detection, and citation type recognition --- and demonstrate the significance of CORWA with analyses from multiple perspectives (\cref{sec:CORWA}). (2) We propose a strong baseline model that automatically tags the CORWA annotation scheme on massive unlabeled related work section texts (\cref{sec:joint_tagger}). (3) We show that citation spans are a better target than citation sentences with two example tasks (\cref{sec:span_vs_sentence}). (4) We discuss a novel framework for human-in-the-loop, iterative, abstractive related work generation (\cref{sec:full_related_work_generation}).

\section{Related Work}
\paragraph{Extractive Related Work Generation.} 
Early related work generation systems employed the extractive summarization approach. \citet{hoang2010towards} pioneered the task, developing  rules to select sentences following a topic hierarchy tree that was assumed to be given as input. \citet{hu2014automatic} 
grouped sentences into topic-biased clusters with PLSA, modeled sentence importance with SVR, and applied a global optimization framework to select sentences. 
\citet{chen2019automatic} selected sentences from papers that co-cited the same cited papers as the target paper in order to cover a minimum Steiner tree constructed from the paper's keywords. \citet{wang2019toc} extracted Cited Text Spans (CTS), the matched text spans in the cited paper that are most related to a given citation. 
However, these extractive approaches aim to maximally cover the citation texts with the extracted sentences, thus mostly ignoring the \textit{reference} type citations that are concise and abstractive (\cref{sec:citation_type}).

\paragraph{Abstractive Related Work Generation.} 
Recently, \citet{xing2020automatic} extend the pointer-generator \cite{see-etal-2017-get} to take two text inputs, allowing them to recover a masked citation sentence given its neighboring context sentences. \citet{ge-etal-2021-baco} encode the citation context, cited paper's abstract, and citation network and train their model with multiple objectives: sentence salience score regression of the cited paper's abstract, functional role classification of the citation sentence, and citation sentence generation. \citet{chen-etal-2021-capturing} propose a relation-aware, multi-document encoder to generate a related work paragraph given a set of cited papers. \citet{luu-etal-2021-explaining} fine-tune GPT2 \cite{radford2019language} on scientific texts and explore several techniques for representing documents, such as using extracted named entities.


All of the works described above focus on the generation aspect, while neglecting dataset collection; their datasets are mostly extracted automatically. Moreover, the datasets are not reused, though they are publicly available, because these works all use slightly different problem definitions, and thus the models are not directly comparable \cite{li2022automatic}. In this work, we focus on collecting a dataset that is widely applicable to various related work generation settings, rather than proposing another incomparable approach.

\section{CORWA Dataset}
\label{sec:CORWA}
In this work, we limit our scope to publications from the NLP domain for ease of automatically extracting the related work section; existing work on related work generation has also focused on NLP in the past. 
We build our dataset on top of the NLP partition of the S2ORC dataset \cite{lo-wang-2020-s2orc}, a large-scale corpus of scientific papers derived from \LaTeX{} source code and PDF files. We extract the related work section by matching the section titles. Because not all papers cited in the extracted related work sections are available in S2ORC, we prioritize annotating related work sections where the majority of their cited papers are available.

\subsection{Annotation Scheme}
Our CORWA dataset decomposes the related work section with three inter-related annotation tasks: discourse tagging, citation span detection, and citation type recognition. 

\subsubsection{Discourse Tagging}
Each sentence in a related work section has a specific role and information source. Some may be general topic or transition sentences; some summarize one or multiple prior works in detail, while others describe the general relationship among prior works at a high level. Our discourse tagging task tags the role of each related work sentence with one of six labels: \{\emph{single\_summ}, \emph{multi\_summ}, \emph{narrative\_cite}, \emph{reflection}, \emph{transition}, \emph{other}\}.

\paragraph{Single Document Summarization.} 
\emph{Single\_summ} refers to sentences that summarize one single cited work in detail. Most typically, this includes sentences with explicit citation marks, as when a work is mentioned for the first time. We also include the following cases: (1) follow-up sentences without explicit citation marks that describe the same paper as a preceding \emph{single\_summ} sentence, and (2) sentences containing multiple citations that heavily focus on one of those works. 

\paragraph{Multi-Document Summarization.} 
\emph{Multi\_summ} refers to sentences that summarize multiple prior works of equal importance. As with \emph{single\_summ}, we include the case of follow-up sentences without explicit citation marks that continue describing the same group of prior works discussed in a preceding \emph{multi\_summ} sentence.

\paragraph{Narrative Citation.}
In contrast to \emph{single\_summ} and \emph{multi\_summ}, narrative citation (\emph{narrative\_cite}) refers to citation sentences that do not summarize specific cited works in detail, but rather convey high-level observations from the authors of the current work. \emph{Narrative\_cite} sentences may contain general statements about the field or task, or the authors' comments on or comparisons of prior works.

\paragraph{Reflection.}
In addition to describing prior works, authors discuss how they relate to the current work, highlighting the authors' novel contributions. These \emph{reflection} sentences focus on the current work, instead of prior works. 

\paragraph{Transition.}
Non-citation sentences in related work sections serve as topic introductions or transitions from one topic to another. We label these supplemental sentences that do not belong to any of the above cases as \emph{transition} sentences.

\paragraph{Other.}
The related work sections in our dataset are extracted automatically using heuristics based on section titles, and there are occasionally some errors in section boundary detection; we label those sentences that are not actually part of the related work section as \emph{other}.

\subsubsection{Citation Span Detection} \label{sec:citation_span}
In order to understand sentences that describe prior work, it is crucial to recognize the token-level mapping between the citation text and the cited paper(s). Our citation span detection task identifies the span of text whose information is directly derived from a specific cited paper. 
For example, if a cited paper is explained with a summary, its citation span covers the entire summary, which may range from part of a sentence to a few consecutive sentences; if a cited paper is mentioned with an explicit citation, but is not described or discussed at all, then the citation span is just the citation mark. 

In constructing the dataset, we find that a single citation rarely spans across paragraph boundaries without a new explicit citation mark, so we require our spans to be bounded by paragraph boundaries. 

\subsubsection{Citation Type Recognition} \label{sec:citation_type}
Our citation type recognition task indicates whether a cited work is discussed in detail or used to illustrate a high-level concept. We label these types of citations as \emph{dominant} and \emph{reference}, respectively.

\paragraph{Dominant.} These citations are discussed in detail, usually via summarization of their content, and are often longer than \textit{reference} citations.

\paragraph{Reference.} These citations are not discussed in detail. They frequently appear in \emph{narrative\_cite} sentences, but may also appear in \emph{single\_summ} and \emph{multi\_summ} sentences when they are not the main focus of the sentence, and thus it is not sufficient to depend on the sentence-level discourse tags to distinguish them. For example, in Figure \ref{fig:CORWA_example}, line 5, the pointer-generator network \cite{see-etal-2017-get} is cited for reference as part of a longer \textit{dominant} citation span. \textit{Reference} citations tend to be more abstractive than \textit{dominant} citations. 

\begin{table*}[t]
\begin{center}
\small
    \begin{tabular}{ l l l l l l l l l }
    \hline
    \textbf{Disc. Label} $(d)$ & \textbf{$n(d)$} & \textbf{$p(d)$} & \textbf{$p(d\given D)$} & \textbf{$p(d\given R)$} &\textbf{$p(D\given d)$} & \textbf{$p(R\given d)$} &\textbf{$p(D, d)$} & \textbf{$p(R, d)$}  \\ \hline
    \emph{single\_summ} & $4255$ & $30.8\%$ & $80.8\%$ & $1.1\%$ & $98.5\%$ & $1.5\%$ & $\textbf{36.9\%}$ & $0.6\%$\\ \hline
    \emph{transition} & $3371$ & $24.4\%$ & $0$ & $0.2\%$ & $12.5\%$ & $87.5\%$ & $0$ &$0.1\%$ \\ \hline
    \emph{narrative\_cite} & $2540$ & $18.4\%$ & $0.4\%$ & $90.2\%$ & $0.4\%$ & $99.6\%$ & $0.2\%$ & $\textbf{48.9\%}$\\ \hline
    \emph{reflection} & $2489$ & $18.0\%$ & $0.1\%$ & $6.1\%$ & $1.5\%$ & $98.5\%$ & $0.1\%$ & $3.3\%$ \\ \hline
    \emph{multi\_summ} & $671$ & $4.8\%$ & $18.7\%$ & $2.5\%$ & $86.4\%$ & $13.6\%$ & $\textbf{8.5\%}$ & $1.3\%$ \\ \hline
    \emph{other} & $510$ & $3.7\%$ & $0$ & $0$ & $0$ & $100.0\%$ & $0$ & $0$\\ \hline
    \end{tabular}
    \caption{Distributions of discourse labels and citation spans in CORWA dataset. $d$: Discourse labels. $D/R$: \textit{Dominant}/\textit{reference} type citation span. $n(D)=3565, n(R)=4228$. 2927 paragraphs in total.} \label{tab:discourse_probablity_full}
    \vspace{-1em}
\end{center}
\end{table*}

\subsection{Annotation Process and Agreement}
Two graduate students from our university's Computer Science Department\footnote{One of them later became the second author of this paper.}, manually annotated 927 related work sections. They first annotated 23 related work sections from scratch, after which we incrementally trained a transformer-based tagging model \cite{vaswani2017attention} (\cref{sec:joint_tagger}) to assist the annotation process, asking the annotators to correct the model's predictions, rather than performing manual annotation from scratch. We split the 362 annotated related work sections from papers published in 2019 and later as our test set and all 565 earlier papers as the training set. 

Since each related work section is labeled by a single annotator, we calculate agreement by sampling 50 related work sections from the test set and asking the other annotator to re-annotate them from scratch\footnote{The disagreements are adjudicated by the first author.}. We obtain strong agreement on all tasks (Cohen's $\kappa$ of 0.824, 0.965 and 0.878 for discourse tagging, citation type recognition, and citation span detection, respectively; citation type recognition and citation span detection are converted to token-level labels for agreement calculation).

The automated, correction-based annotation process is much faster than annotating from scratch and allows us to collect a much larger annotated dataset. As a trade-off, the annotations may be biased by the model's predictions if the annotators fail to notice any incorrect predictions. This may explain why our model performance reported in \cref{sec:joint_tagger_experiment} is higher than the inter-annotator agreement. 

\subsection{Analysis of CORWA}
The tasks of discourse tagging, citation span detection, and citation type recognition, capture distinct but overlapping perspectives of information. 

\begin{figure}
\centering
    \includegraphics[width=0.4\textwidth, height=0.3\textwidth]{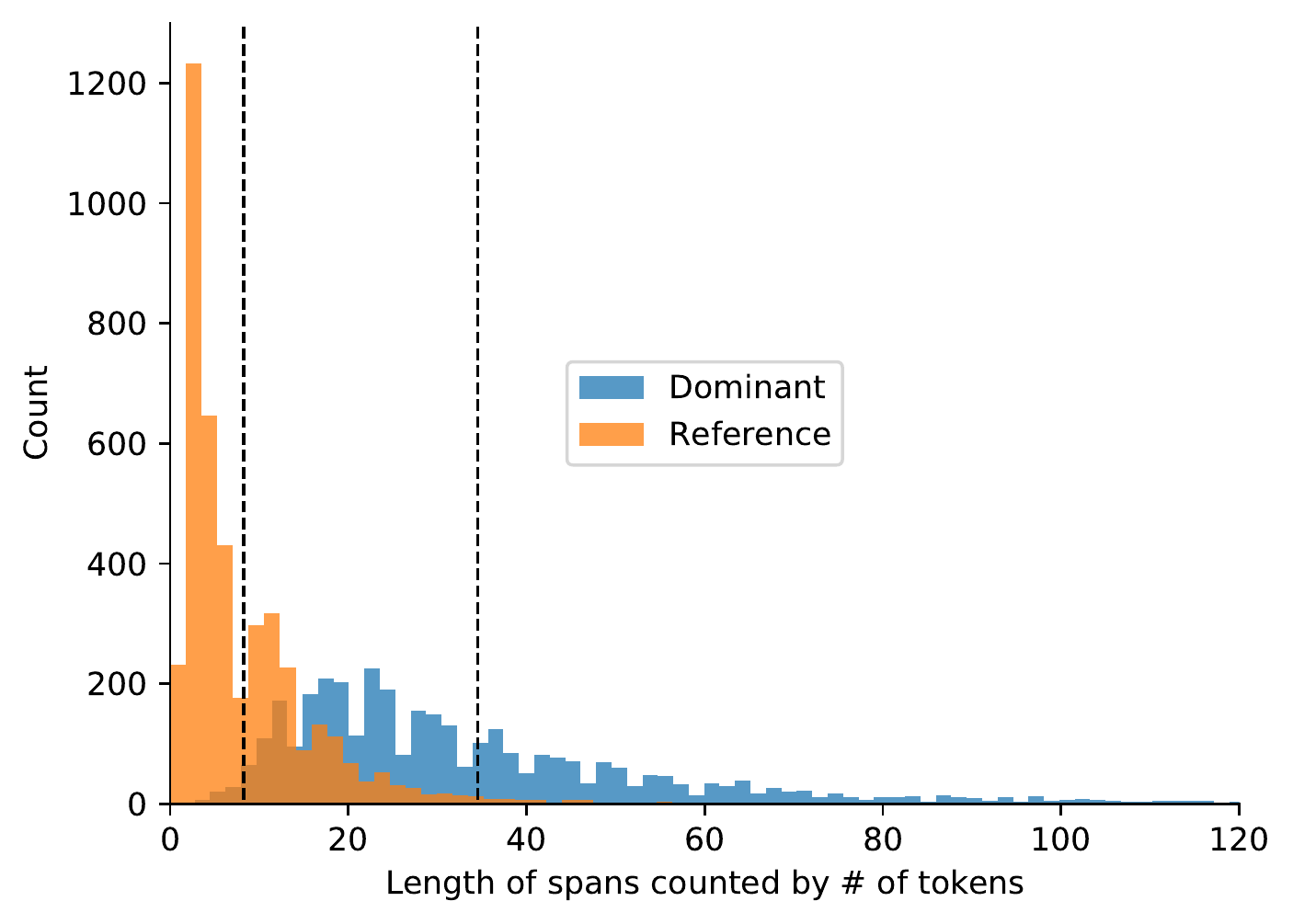}
     \vspace{-0.5em}
  \caption{Histogram of the length of \textit{dominant} and \textit{reference}-type citation spans, excluding citation marks. The dashed vertical lines are the means of \textit{dominant} and \textit{reference} span lengths, 34.5 and 8.2, respectively.} 
  \label{fig:span_lens}
  \vspace{-1em}
\end{figure}

\subsubsection{Relations among CORWA Subtasks}
We investigate the relationships among the CORWA subtasks by calculating the co-occurrence distributions of discourse labels and citation span types. A citation span is considered \textit{dominant} if it contains any \textit{dominant} citations, and \textit{reference} otherwise. Figure \ref{fig:span_lens} shows that \textit{dominant}-type spans (average of 34.5 tokens) are significantly longer than \textit{reference}-type spans (average of 8.2 tokens).

Table \ref{tab:discourse_probablity_full} shows the count of each discourse label, the conditional probability and the joint probability of discourse labels and citation span types. \emph{Single\_summ} with \textit{dominant} span, \emph{multi\_summ} with \textit{dominant} span, and \emph{narrative\_cite} with \textit{reference} span are the most frequent combinations
. These observations make intuitive sense, since \textit{dominant}-type spans describe cited papers in detail, often taking the form of a summary, while \textit{reference}-type spans are highly abstracted, making them more likely to be mixed into \textit{narrative}-type sentences that discuss high-level ideas, often encompassing multiple cited papers. This difference is analogous to \textit{informative} versus \textit{indicative} summaries, where the former serves as a surrogate for the document, and the latter characterizes what the document is about \cite{kan-etal-2001-applying}.

\subsubsection{Related Work Writing Styles}

\begin{figure}
\centering
    \includegraphics[width=0.4\textwidth, height=0.35\textwidth]{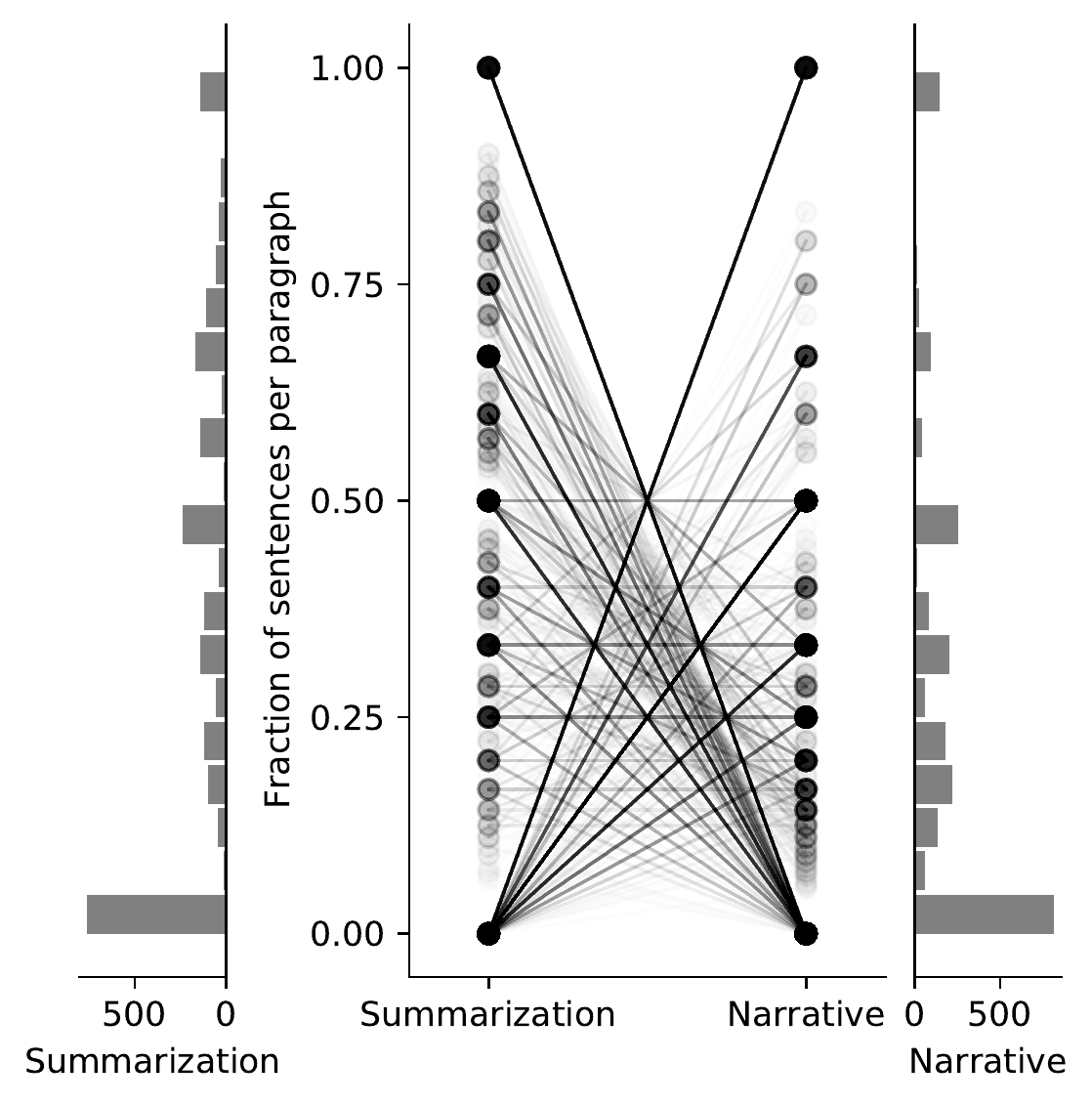}
     \vspace{-0.5em}
  \caption{Parallel plot of the proportion of \textit{summarization} and \textit{narrative} sentences in each paragraph. Paragraphs with neither type of sentences are excluded.} 
  \label{fig:Summ_Narrative_parallel_histogram}
  \vspace{-1.5em}
\end{figure}

\paragraph{Integrative or Descriptive?}
As \citet{khoo2011analysis} note, authors may describe the same cited paper in two different styles: descriptive, which explicitly summarizes the cited paper, or integrative, which describes and comments on the cited paper in a narrative form. We examine the ratio of \emph{summarization} (both \emph{single\_summ} and \emph{multi\_summ}) and \emph{narrative} sentences (\emph{narrative\_cite}) in related work paragraphs (Figure \ref{fig:Summ_Narrative_parallel_histogram}). The CORWA discourse labels capture writing style differences among papers: 34.6\% of related work section paragraphs only contain \textit{summarization} sentences, resembling Khoo et al.'s descriptive literature review, while 32.1\% of paragraphs contain only \textit{narrative} sentences, resembling an integrative literature review. Interestingly, 33.3\% of paragraphs mix both styles and are neither purely descriptive nor purely integrative.

\paragraph{Frequent Discourse Label Subsequences.}
Scientific discourse is used by paper authors to promote their ideas \cite{li2021scientific}. We analyze the patterns of CORWA discourse labels to uncover how authors promote their ideas using a mix of sentence types. We apply the rule-based PrefixSpan \cite{han2001prefixspan} and Gap-Bide \cite{li2008efficiently} algorithms to extract frequent discourse label subsequences. We identify six typical subsequences, shown in Supplementary Tables \ref{tab:discourse_subsequence} and \ref{tab:discourse_subsequence_countinue}. For example, the pattern of \emph{single\_summ} followed by \emph{reflection} compares the cited paper to the current work, usually without directly criticizing the cited paper, while \textit{single\_summ} followed by \textit{transition} is the more impersonal pattern for criticism of a cited paper, where authors tend to avoid direct comparison with the current work. 




\section{Joint Related Work Tagger}

\begin{figure}
\centering
    \includegraphics[width=0.4\textwidth, height=0.3\textwidth]{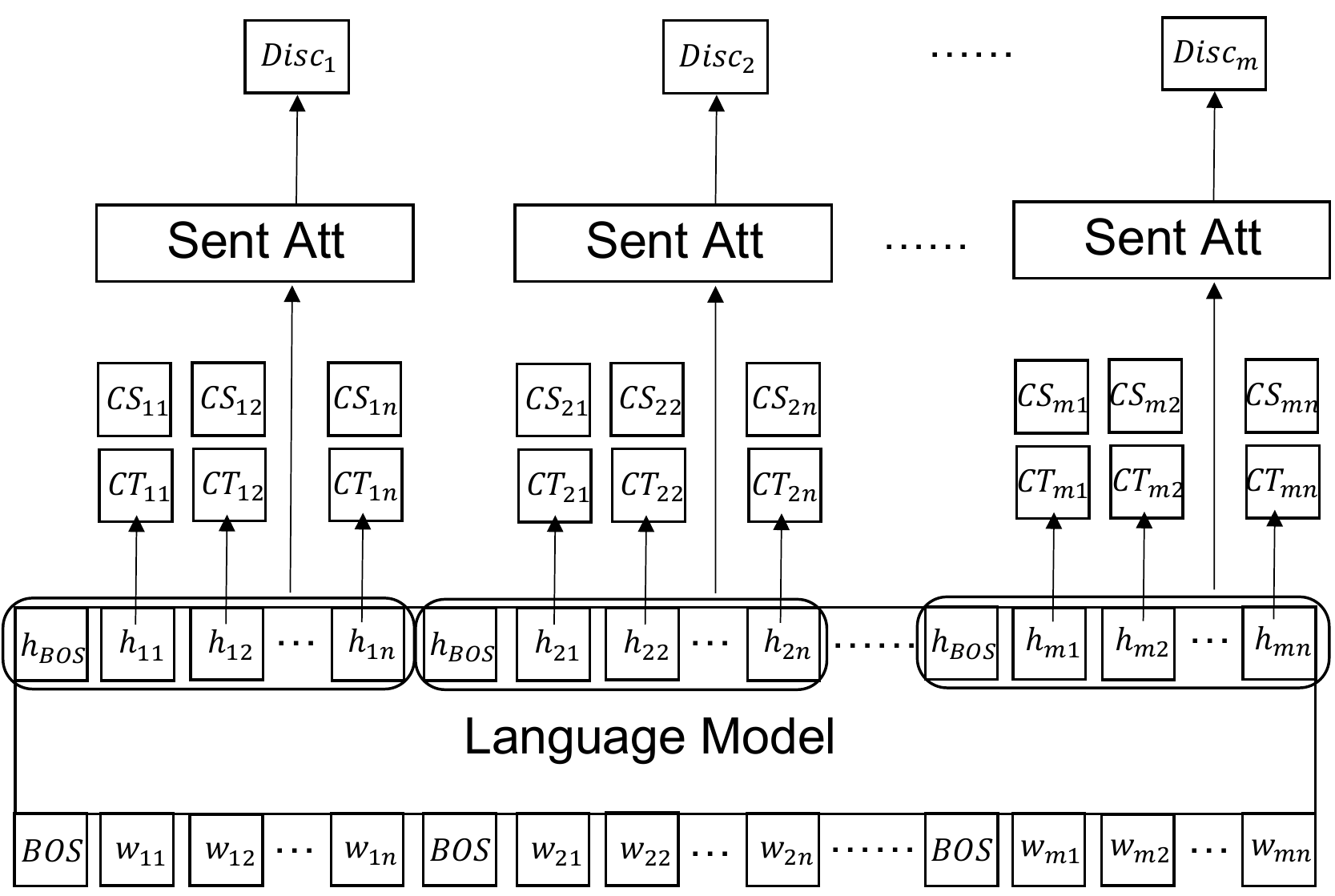}
     \vspace{-0.5em}
  \caption{The architecture of our joint related work tagger, which performs discourse tagging (Disc), citation type recognition (CT), and citation span detection (CS).} 
  \label{fig:joint_tagger}
  \vspace{-1.5em}
\end{figure}

\label{sec:joint_tagger}
To help propagate our CORWA annotations to massive unlabeled related work sections, we build a joint related work tagger baseline\footnote{\url{https://github.com/jacklxc/CORWA}} that is trained on the three annotation tasks, discourse tagging, citation span detection, and citation type recognition, via multi-task learning \cite{caruana1997multitask}. 

\subsection{Model Design}
Figure \ref{fig:joint_tagger} shows the model architecture of our joint related work tagger. We encode related work sections using a transformer-encoder \cite{vaswani2017attention} paragraph by paragraph, as we enforce the independence of paragraphs in CORWA citation span annotations. We decode citation span labels and citation type labels token by token, while our discourse tagging task uses the paragraph-level sentence tagging mechanism proposed by \citet{li2020paragraph}. Because the three sub-tasks of CORWA are inter-related, we use multi-task learning to jointly train the tagger by sharing the encoder across tasks.

\subsubsection{Paragraph Encoder} 
\label{sec:paragraph_encoder}

We experiment with several pre-trained transformer-encoders \cite{devlin-etal-2019-bert, beltagy-etal-2019-scibert, liu2019roberta, Beltagy2020Longformer}, and eventually focus on SciBERT \cite{beltagy-etal-2019-scibert}, which is a variant of the BERT model \cite{devlin-etal-2019-bert} that is trained on a scientific corpus with domain-specific tokenization schemes, including NLP papers.


\subsubsection{Task-specific Decoders}
\label{sec:joint_tagger_detail}
\paragraph{Citation Span Detection \& Citation Type Recognition.}
We use the $BIO2$ tagging scheme \cite{sang1999representing} for the citation span detection and citation type recognition tasks; we use \textit{B}, \textit{I}, \textit{O} for citation span detection and five labels --- \textit{B-Dominant}, \textit{I-Dominant}, \textit{B-Reference}, \textit{I-Reference}, and \textit{O} --- for citation type recognition. We use a two-layer feed-forward network to decode the encoded paragraph-level token embeddings to the output sequence of $BIO2$ tags.

\paragraph{Discourse Tagging.}
We apply \citet{li2020paragraph}'s paragraph-level sentence tagging approach for the discourse labels: a simple attention mechanism is used to aggregate token embeddings, sentence by sentence, into sentence encodings, before decoding the sentence encodings into discourse labels using a two-layer multi-layer feed-forward network.

\subsubsection{Multi-task Learning}
We use cross-entropy loss on all three CORWA sub-tasks. We balance the relative importance of the sub-tasks by taking a weighted sum of the sub-task losses of discourse tagging, citation span detection, and citation type recognition \{$L_{d}, L_{s}, L_{t}$\}:

\begin{equation}
\begin{aligned}
L &= \gamma_{d} L_{d} + \gamma_{s} L_{s} + \gamma_{t} L_{t}
\end{aligned}
\end{equation}

where \{$\gamma_{d}, \gamma_{s}, \gamma_{t}$\} are tuned hyper-parameters; their values are given in Table \ref{tab:hyper_parameter}.

\begin{table}[t]
\begin{center}
\small
    \begin{tabular}{  l  l  l  l  }
    \hline
    \textbf{Model} & \textbf{Disc} & \textbf{CT} & \textbf{CS} \\ \hline
    SciBERT & $0.898$ & $0.959$ & $0.930$ \\ 
    + Distant Dataset & $\textbf{0.908}$ & $\textbf{0.963}$ & $\textbf{0.933}$ \\ \hline
    \end{tabular}
    \vspace{-0.5em}
    \caption{Test set micro-F1 scores of the SciBERT-based joint related work tagger, with and without training on distantly labeled data, on the discourse tagging (Disc), citation type recognition (CT), and citation span detection (CS) tasks.} \label{tab:joint_tagger_performance}
    \vspace{-1.5em}
\end{center}
\end{table}

\begin{table}[t]
\begin{center}
\small
    \begin{tabular}{  l  l }
    \hline
    \textbf{Parameter Name} & \textbf{Value} \\ \hline
    Encoder Learning Rate & $10^{-5}$ \\ \hline
    Decoder Learning Rate & $5 \times 10^{-6}$ \\ \hline
    Dropout & $0$ \\ \hline
    Epoch & $15$ \\ \hline
    Batch Size & $1$ \\ \hline
    Steps per Update & $10$ \\ \hline
    $\gamma_{d}$ & $1$ \\ \hline
    $\gamma_{t}$ & $3$ \\ \hline
    $\gamma_{s}$ & $1.75$ \\ \hline
    \end{tabular}
    \vspace{-0.5em}
    \caption{Hyper-parameters of our best joint related work tagger (SciBERT + Distant Dataset).} \label{tab:hyper_parameter}
    \vspace{-1.5em}
\end{center}
\end{table}

\subsection{Experiments} \label{sec:joint_tagger_experiment}
 We perform five-fold cross-validation to tune the model hyper-parameters. Table \ref{tab:joint_tagger_performance} shows the strong performance of the model\footnote{Supplementary Table \ref{tab:full_joint_tagger_performance} shows the full cross-validation and test performances.}. We use the joint related work tagger to automatically label the unannotated 11,465 related work sections remaining in the S2ORC NLP partition and then use this distantly-supervised data to further boost the model's performance. For the citation span detection and citation type recognition tasks, we use a token-level F1 score. Our final, distantly-supervised joint related work tagger achieves more than 0.9 test F1 on all three tasks, indicating the high quality of the model's predictions. This model can be used to propagate our labels on the unannotated related work sections to create a very large training set for future work. 


\section{Spans as an Alternative to Sentences}
\label{sec:span_vs_sentence}
We argue that the citation spans annotated in CORWA are a better alternative to the citation sentences that have previously been used for the tasks of ROUGE-based retrieval and citation text generation.

\begin{figure}
\centering
    \includegraphics[width=0.4\textwidth, height=0.3\textwidth]{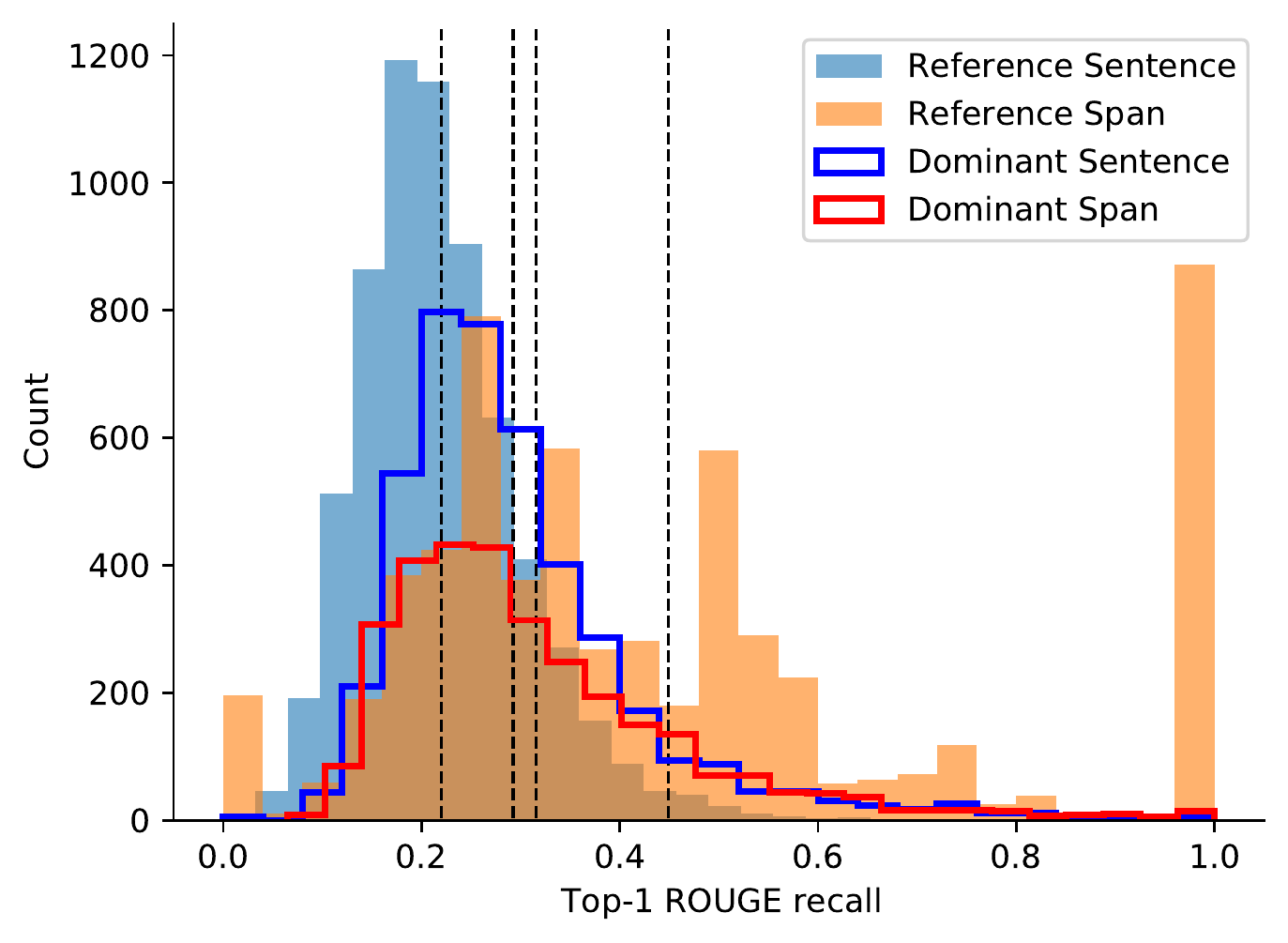}
     \vspace{-0.5em}
  \caption{Histogram of top-1 ROUGE recall scores of retrieved sentences from cited papers using different queries. The dashed vertical lines are the means of reference sentence (0.220), dominant sentence (0.293), dominant span (0.316), and reference spans (0.449).} 
  \label{fig:ROUGE-retrieval}
  \vspace{-1.5em}
\end{figure}

\subsection{Queries for Relevant Sentence Retrieval}
\label{sec:retrieval}
Citations focus on a small portion of the content in cited papers, and this focus is not explicitly recorded in the citation network. A popular approach for determining relevant sentences retrieves sentences from the cited papers by comparing the similarity between the gold citation sentence and candidate sentences in the cited paper \cite{cao2015ranking, yasunaga2017graph, yasunaga2019scisummnet, ge-etal-2021-baco}. Figure \ref{fig:ROUGE-retrieval} compares the distribution of the top-1 average of ROUGE-1 and ROUGE-2 recall scores \cite{lin2004rouge} of retrieved sentences from cited papers using citation spans with those using citation sentences\footnote{Only papers included in S2ORC dataset are considered.}. There is no significant difference between the average ROUGE scores of \textit{dominant} spans and sentences containing \textit{dominant} citations, which is reasonable because \textit{dominant} spans are often full sentences anyway. In contrast, the average score of \textit{reference} spans is significantly higher than that of sentences containing \textit{reference}-type citations; \textit{reference} spans are shorter and contain highly concentrated key information derived from their cited papers. Thus, using CORWA citation spans as queries for ROUGE-based cited sentence retrieval is superior for \textit{reference}-type citations and comparable for \textit{dominant}-type citations.

\begin{figure}
\centering
    \includegraphics[width=0.4\textwidth, height=0.3\textwidth]{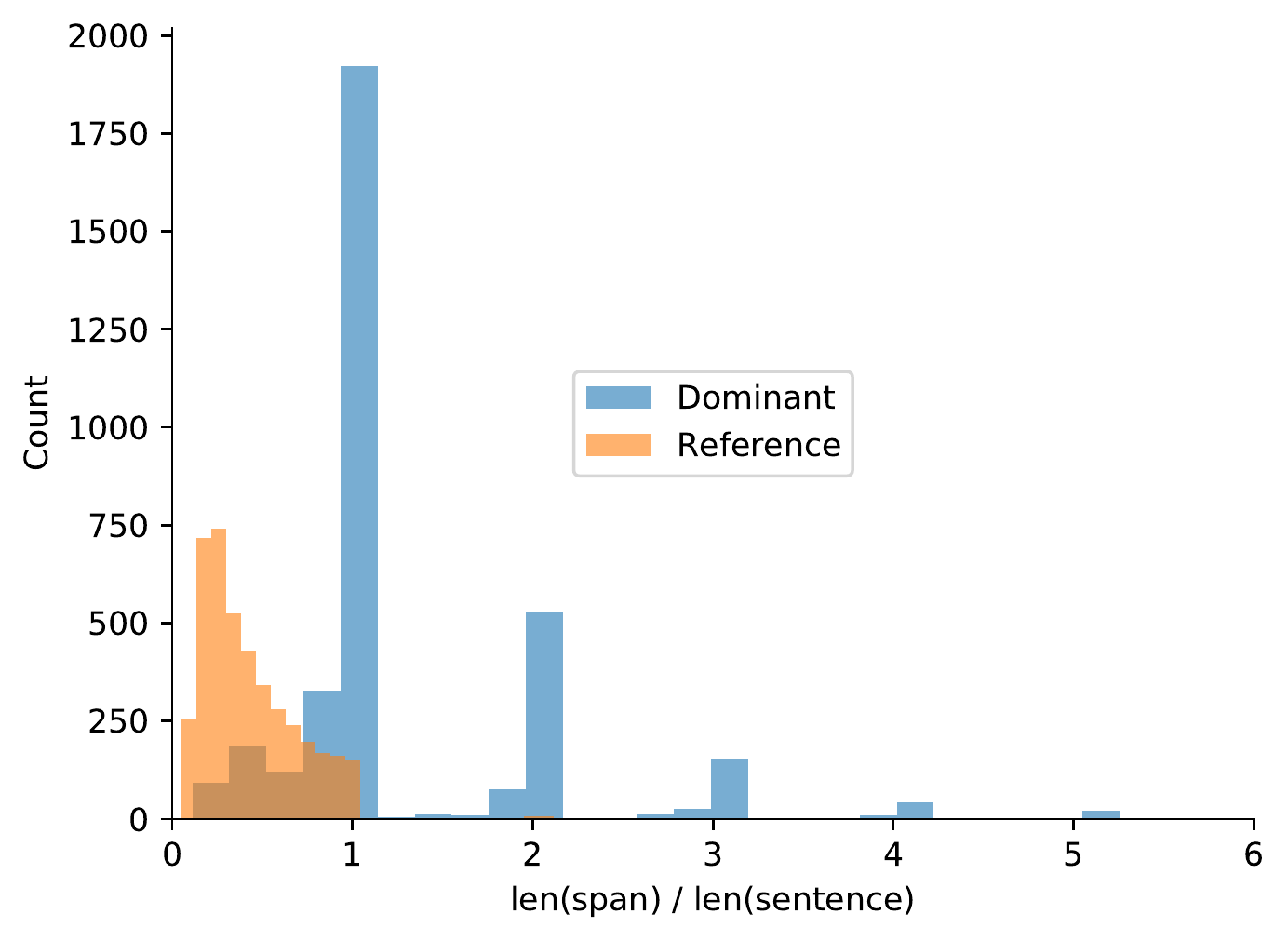}
     \vspace{-0.5em}
  \caption{Histogram of the ratio of between the lengths of \textit{dominant} and \textit{reference} type citation spans and the corresponding citation sentences. None of the reference spans are longer than one sentence. 27.7\%, 46.6\%, and 25.7\% of \textit{dominant} spans are shorter than, equal to, or longer than one sentence, respectively.} 
  \label{fig:span_sentence_lens}
  \vspace{-1.5em}
\end{figure}

\subsection{Span-based Related Work Generation} \label{sec:span_geneartion}
Existing neural network-based, abstractive related work generation systems generate citation sentences given the surrounding context sentences \cite{xing2020automatic, ge-etal-2021-baco, luu-etal-2021-explaining} or generate entire paragraphs containing multiple citations \cite{chen-etal-2021-capturing}. These task settings neglect the fact that the citation text corresponding to a cited paper is not necessarily in the form of a sentence, but could be a portion of a sentence or a block of multiple sentences. Our span-based annotation scheme identifies the citation tokens that are directly derived from the cited papers. 

As Figure \ref{fig:span_sentence_lens} shows, \textit{reference} spans are not full sentences, while \textit{dominant} spans can cover multiple sentences. For \textit{reference}-type citations, using a full sentence as the generation target includes potentially unrelated tokens outside the citation span that do not refer to the cited paper. For \textit{dominant}-type citations, using a single sentence as the generation target can result in 1) information loss when not all sentences describing the cited paper are included in the target, and the model never learns to generate them, or 2) information leak when sentences that actually describe the cited paper are used as context sentences instead of target sentences. Thus, we propose a span-level citation text generation task and present a pilot study using a Longformer-Encoder-Decoder (LED) \cite{Beltagy2020Longformer} baseline model.

\subsubsection{Experimental Setting} \label{sec:span_geneartion_setting}
The common Transformer-based language models \cite{devlin-etal-2019-bert, liu2019roberta, lewis-etal-2020-bart, 2020t5} have a limited input window size (typically 512 or 1024 tokens), which presents a major challenge for tasks like related work generation that use multiple long documents as inputs. LED \cite{Beltagy2020Longformer} addresses this challenge by using a local self-attention mechanism, rather than global self-attention, handling in input context windows of up to 16k tokens. We present an LED-based baseline model for the citation span generation task. 

We first pretrain the LED-base model 
on the masked language modeling (MLM) task \cite{devlin-etal-2019-bert} using related work sections from S2ORC papers in the computer science domain, as well as on the cross-document language modeling (CDLM) task \cite{caciularu2021cdlm}, which aligns masked citation sentences with their context sentences and the full text of their cited papers. We further pretrain the LED encoder with the three CORWA sub-tasks (Supplementary Table \ref{tab:full_joint_tagger_performance}). All pretraining strictly excludes the texts from test set. 

For the citation span generation task, we input the concatenation of \{the target paper's introduction (following \citet{luu-etal-2021-explaining}), the partial related work paragraph excluding the target citation span, and the concatenation of \{explicit citation mark, title, and abstract\} of each cited paper in the target span\footnote{We indicate whether the target span is \textit{dominant} or \textit{reference} type, as well as the type of each citation in the span.}\}; the generation target is the ground truth citation span from CORWA. We provide the explicit citation mark (e.g. Devlin et al., 2018) because it is simple to extract but cannot be inferred from the paper text alone. Just as a human reader may remember the content of the frequently cited papers or the research topics of frequently cited authors, so the citation mark tokens may carry information about the cited paper and its authors. 

In addition to the CORWA training set, we use the distantly supervised labels predicted by our joint related work tagger (\cref{sec:joint_tagger_experiment}) for training. We use the default hyper-parameters of the Huggingface LED implementation \cite{wolf-etal-2020-transformers}.

\subsubsection{Experimental Results}

\begin{table*}[t]
\begin{center}
\small
    \begin{tabular}{  l | l  l  l | l l l}
    \hline
    & \multicolumn{3}{l}{\textit{Dominant}} & \multicolumn{3}{l}{\textit{Reference}} \\
    \textbf{Models} & \textbf{R-1} & \textbf{R-2} & \textbf{R-L} & \textbf{R-1} & \textbf{R-2} & \textbf{R-L} \\ \hline
    LED-base w/o pretrain & $0.220$ & $0.060$ & $0.183$ & $0.228$ & $0.091$ & $0.223$ \\ 
    LED-base Span  & $0.230$ & $0.062$ & $0.186$ & $0.244$ & $0.107$ & $0.240$ \\ \hline
    LED-base Sentence  & $0.244$ & $0.075$ & $0.202$ & $0.193$ & $0.050$ & $0.151$ \\ \hline
    \end{tabular}
    \vspace{-0.5em}
    \caption{Performance of citation span/sentence generation using LED-base \cite{Beltagy2020Longformer}. Citation marks are excluded from the scores since they are trivial to generate and bring up the scores unintentionally. Note that the performance of span/sentence generations are NOT directly comparable due to different generation targets.} \label{tab:span_generation}
    \vspace{-1.5em}
\end{center}
\end{table*}

As Table \ref{tab:span_generation} shows, the ROUGE scores of our LED-base models for citation span/sentence generation are similar to previous sentence-level citation text generation models \cite{xing2020automatic, ge-etal-2021-baco}, and our pretraining improves the citation span generation performance. Compared to sentence-level generation, span-level generation has lower scores for \textit{dominant} citations, but higher scores for \textit{reference} citations. However, because the span- and sentence-level tasks have different generation targets, their scores cannot be directly compared.

We perform a human evaluation following the setting of \citet{xing2020automatic, ge-etal-2021-baco}. We sample 15 instances each for \textit{dominant} and \textit{reference} citations and compare their corresponding span- and sentence-based generation outputs, as well as the gold spans from the original related work sections. Each citation text is rated by three NLP graduate students who are fluent in English on a 1 (very poor) to 5 (excellent) point scale, with respect to four aspects: \textit{fluency} (whether a citation span/sentence is fluent), \textit{relevance} (whether a citation span/sentence is relevant to the cited paper(s)), \textit{coherence} (whether a citation span/sentence is coherent within its context), and \textit{overall quality}.

\begin{table}[t]
\begin{center}
\small
    \begin{tabular}{  l  l  l  l  l }
    \hline
     & Flu.& Rel. & Coh. & Overall \\ \hline
     \textbf{\textit{Dominant}} & & & & \\
     Gold Span & $4.61$ & $3.53$ & $4.17$ & $3.64$ \\
     Span & $\textbf{4.92}$ & $\textbf{4.07}$ & $\textbf{4.20}$ & $3.99$\\
     Sentence & $4.83$ & $4.03$ & $4.17$ & $\textbf{4.02}$\\
    \hline 
     \textbf{\textit{Reference}} & & & & \\
     Gold Span & $\textbf{4.87}$ & $4.04$ & $4.18$ & $\textbf{4.00}$\\
     Span & $4.68$ & $\textbf{4.24}$ & $\textbf{4.26}$ & $3.96$\\
     Sentence & $4.86$ & $3.64$ & $4.09$ & $3.70$\\
    \hline
    \end{tabular}
    \vspace{-0.5em}
    \caption{Average fluency, relevance, coherence and overall scores, rated by human judges.} \label{tab:human_evaluation}
    \vspace{-1.5em}
\end{center}
\end{table}

Table \ref{tab:human_evaluation} shows human evaluation results, with moderate inter-annotator agreement (Kendall's $\tau$ of 0.298, 0.205, and 0.172 among three annotators). All citation texts are judged to be highly fluent. 

Interestingly, in previous studies \cite{xing2020automatic, ge-etal-2021-baco} the scores of gold sentences are higher than those of generated texts, but our gold spans have a significantly lower relevance scores than the generated spans. This is likely because the gold spans contain information derived from the body sections of the cited papers, which are not provided to either the models or to the human judges. As a result, some gold spans appear to be irrelevant to the human judges, echoing our earlier finding in \cref{sec:retrieval} that citation spans contain more focused information. This observation also suggests that gold citation spans are not necessarily the best target for all task settings. 

We also see that, while \textit{dominant} sentences and spans receive similar scores, the \textit{reference} sentences have lower relevance scores than the spans. This result makes sense because \textit{reference} citation spans are short and focused, so the full sentences include tokens unrelated to the cited paper(s). Overall, the generated spans are rated slightly higher than the generated sentences by the human judges, confirming that span-level citation text generation is preferable to sentence-level generation.

\section{Toward Full Related Work Generation}
\label{sec:full_related_work_generation}
Existing extractive related work generation systems \cite{hoang2010towards, hu2014automatic, chen2019automatic, wang2019toc} select sentences from the target paper and/or the cited papers, which can be concatenated to form a full related work section; neural network-based, abstractive related work generation systems generate individual citation sentences \cite{xing2020automatic, ge-etal-2021-baco, luu-etal-2021-explaining} or paragraphs \cite{chen-etal-2021-capturing}. However, none of these prior works address the ordering of the extracted/generated sentences or the grouping of sentences into paragraphs, nor are they able to produce rhetorical sentences to smooth the transitions between citations. No prior work bridges the gap from generating individual citation texts to generating a full related work section. 

We suggest a bottom-up, iterative approach to generate full related work sections. The process would begin with generating citation spans under the settings proposed in \cref{sec:span_geneartion}. Then, multiple generated citation spans would be aggregated and rewritten into citation text blocks in either the \textit{summarization} or \textit{narrative} style. These blocks would be further aggregated and rewritten into paragraphs by generating \textit{transition} and \textit{reflection} sentences. 

Generating and rewriting in this pipeline fashion has the following benefits: (1) It mitigates the practical issue of computational resource limitations, given that state-of-the-art models do not perform well on long text generation. (2) The auxiliary inputs, such as citation functions or discourse tags, may vary for each stage of generation. (3) As a practical system to assist researchers, it is crucial to allow user involvement in the iterative generation process. Due to the large search space, consisting of multiple valid related work section candidates with different writing styles, it is extremely challenging to precisely generate a satisfying text with a one-shot, end-to-end system. A human-in-the-loop approach allows the user to significantly prune the search space and simultaneously reduces the error-propagation issue caused by the pipeline design.


\subsection{Other Related Tasks}
\subsubsection{Scientific Document Understanding}
Besides summarization, scientific document understanding also plays an important role in related work generation.

\paragraph{Citation Analysis.}
Citations are the core of related work sections. There has been a line of research on citation analysis, including citation function \cite{teufel2006automatic, dong2011ensemble, jurgens2018measuring, tuarob2019automatic}, citation intent \cite{cohan-etal-2019-structural, lauscher2021multicite}, 
citation sentiment \cite{athar2011sentiment, athar2012context, ravi2018article, vyas2020article}, etc. These studies annotate citations with different labeling schemes to study the various usages and purposes of citations.

\paragraph{Discourse Analysis.} 
Scientific discourse analysis studies the rhetorical components of clauses, sentences, or text spans that are not limited to citations, uncovering how authors persuade expert readers with their claims. There is a significant amount of prior work proposing discourse schemes and developing models for discourse tagging for scientific articles \cite{teufel1999discourse, teufel2002summarizing, hirohata2008identifying, liakata2010zones, liakata2012automatic, guo2010identifying, de2012epistemic, burns2016automated, PubMed-RCT, huang-etal-2020-coda, li2021scientific}. 

Our CORWA discourse tagging task focuses on distinguishing the source of the information in each related work sentence, which is complementary to the discourse tagging work listed above. 

\subsubsection{Cited Text Span}
\citet{aburaed-etal-2020-multi} extend \citet{hoang2010towards}'s RWSData dataset by annotating the Cited Text Span (CTS) \cite{wang2019toc}. They annotate the specific sentences in cited papers that each citation in the target paper is based on. For each cited paper, they further collect a set of papers that co-cite this cited paper. \citet{jaidka2018insights, jaidka2019cl} propose the CL-Scisumm shared task, which includes identifying the CTS in reference papers for each citation instance. This shared task provides a valuable dataset for the precise generation of citation texts from a CTS, in contrast to most recent work, which uses the cited paper's abstract or introduction.

\subsubsection{Studies of Literature Reviews}
From an information studies perspective, \citet{khoo2011analysis} largely classify literature reviews into two styles: integrative and descriptive. Descriptive literature reviews summarize individual studies and provide detailed information on each, such as methods, results, and interpretation; integrative literature reviews provide fewer details of individual studies, instead focusing on synthesizing ideas and results extracted from these papers. \citet{jaidka2010imitating, jaidka2011literature, jaidka2013literature} analyze the properties of these two types of literature reviews.

\section{Conclusion}
We present the CORWA dataset of three inter-related annotation tasks: discourse tagging, citation span detection, and citation type recognition. We demonstrate the significance of CORWA with analyses from multiple perspectives, such as writing style and discourse patterns. We propose a strong baseline model that can automatically propagate the CORWA annotation scheme to massive unlabeled related work sections. Furthermore, we show that citation spans are a better alternative to citation sentences for both the relevant sentence retrieval and citation generation tasks. Finally, we discuss a novel framework for human-in-the-loop iterative abstractive related work generation.

\clearpage

\bibliography{anthology,custom}

\begin{thebibliography}{60}
\expandafter\ifx\csname natexlab\endcsname\relax\def\natexlab#1{#1}\fi

\bibitem[{AbuRa{'}ed et~al.(2020)AbuRa{'}ed, Saggion, and
  Chiruzzo}]{aburaed-etal-2020-multi}
Ahmed AbuRa{'}ed, Horacio Saggion, and Luis Chiruzzo. 2020.
\newblock \href {https://aclanthology.org/2020.lrec-1.824} {A multi-level
  annotated corpus of scientific papers for scientific document summarization
  and cross-document relation discovery}.
\newblock In \emph{Proceedings of the 12th Language Resources and Evaluation
  Conference}, pages 6672--6679, Marseille, France. European Language Resources
  Association.

\bibitem[{Athar(2011)}]{athar2011sentiment}
Awais Athar. 2011.
\newblock Sentiment analysis of citations using sentence structure-based
  features.
\newblock In \emph{Proceedings of the ACL 2011 student session}, pages 81--87.

\bibitem[{Athar and Teufel(2012)}]{athar2012context}
Awais Athar and Simone Teufel. 2012.
\newblock Context-enhanced citation sentiment detection.
\newblock In \emph{Proceedings of the 2012 conference of the North American
  chapter of the Association for Computational Linguistics: Human language
  technologies}, pages 597--601.

\bibitem[{Beltagy et~al.(2019)Beltagy, Lo, and
  Cohan}]{beltagy-etal-2019-scibert}
Iz~Beltagy, Kyle Lo, and Arman Cohan. 2019.
\newblock \href {https://doi.org/10.18653/v1/D19-1371} {{S}ci{BERT}: A
  pretrained language model for scientific text}.
\newblock In \emph{Proceedings of the 2019 Conference on Empirical Methods in
  Natural Language Processing and the 9th International Joint Conference on
  Natural Language Processing (EMNLP-IJCNLP)}, pages 3615--3620, Hong Kong,
  China. Association for Computational Linguistics.

\bibitem[{Beltagy et~al.(2020)Beltagy, Peters, and
  Cohan}]{Beltagy2020Longformer}
Iz~Beltagy, Matthew~E. Peters, and Arman Cohan. 2020.
\newblock Longformer: The long-document transformer.
\newblock \emph{arXiv:2004.05150}.

\bibitem[{Burns et~al.(2016)Burns, Dasigi, de~Waard, and
  Hovy}]{burns2016automated}
Gully~APC Burns, Pradeep Dasigi, Anita de~Waard, and Eduard~H Hovy. 2016.
\newblock Automated detection of discourse segment and experimental types from
  the text of cancer pathway results sections.
\newblock \emph{Database}, 2016.

\bibitem[{Caciularu et~al.(2021)Caciularu, Cohan, Beltagy, Peters, Cattan, and
  Dagan}]{caciularu2021cdlm}
Avi Caciularu, Arman Cohan, Iz~Beltagy, Matthew~E Peters, Arie Cattan, and Ido
  Dagan. 2021.
\newblock Cdlm: Cross-document language modeling.
\newblock In \emph{Findings of the Association for Computational Linguistics:
  EMNLP 2021}, pages 2648--2662.

\bibitem[{Cao et~al.(2015)Cao, Wei, Dong, Li, and Zhou}]{cao2015ranking}
Ziqiang Cao, Furu Wei, Li~Dong, Sujian Li, and Ming Zhou. 2015.
\newblock Ranking with recursive neural networks and its application to
  multi-document summarization.
\newblock In \emph{Proceedings of the AAAI Conference on Artificial
  Intelligence}, volume~29.

\bibitem[{Caruana(1997)}]{caruana1997multitask}
Rich Caruana. 1997.
\newblock Multitask learning.
\newblock \emph{Machine learning}, 28(1):41--75.

\bibitem[{Chen and Zhuge(2019)}]{chen2019automatic}
Jingqiang Chen and Hai Zhuge. 2019.
\newblock Automatic generation of related work through summarizing citations.
\newblock \emph{Concurrency and Computation: Practice and Experience},
  31(3):e4261.

\bibitem[{Chen et~al.(2021)Chen, Alamro, Li, Gao, Zhang, Zhao, and
  Yan}]{chen-etal-2021-capturing}
Xiuying Chen, Hind Alamro, Mingzhe Li, Shen Gao, Xiangliang Zhang, Dongyan
  Zhao, and Rui Yan. 2021.
\newblock \href {https://doi.org/10.18653/v1/2021.acl-long.473} {Capturing
  relations between scientific papers: An abstractive model for related work
  section generation}.
\newblock In \emph{Proceedings of the 59th Annual Meeting of the Association
  for Computational Linguistics and the 11th International Joint Conference on
  Natural Language Processing (Volume 1: Long Papers)}, pages 6068--6077,
  Online. Association for Computational Linguistics.

\bibitem[{Cohan et~al.(2019)Cohan, Ammar, van Zuylen, and
  Cady}]{cohan-etal-2019-structural}
Arman Cohan, Waleed Ammar, Madeleine van Zuylen, and Field Cady. 2019.
\newblock \href {https://doi.org/10.18653/v1/N19-1361} {Structural scaffolds
  for citation intent classification in scientific publications}.
\newblock In \emph{Proceedings of the 2019 Conference of the North {A}merican
  Chapter of the Association for Computational Linguistics: Human Language
  Technologies, Volume 1 (Long and Short Papers)}, pages 3586--3596,
  Minneapolis, Minnesota. Association for Computational Linguistics.

\bibitem[{De~Waard and Maat(2012)}]{de2012epistemic}
Anita De~Waard and Henk~Pander Maat. 2012.
\newblock Epistemic modality and knowledge attribution in scientific discourse:
  A taxonomy of types and overview of features.
\newblock In \emph{Proceedings of the Workshop on Detecting Structure in
  Scholarly Discourse}, pages 47--55. Association for Computational
  Linguistics.

\bibitem[{Dernoncourt and Lee(2017)}]{PubMed-RCT}
Franck Dernoncourt and Ji~Young Lee. 2017.
\newblock \href {http://arxiv.org/abs/1710.06071} {Pubmed 200k {RCT:} a dataset
  for sequential sentence classification in medical abstracts}.
\newblock \emph{CoRR}, abs/1710.06071.

\bibitem[{Devlin et~al.(2019)Devlin, Chang, Lee, and
  Toutanova}]{devlin-etal-2019-bert}
Jacob Devlin, Ming-Wei Chang, Kenton Lee, and Kristina Toutanova. 2019.
\newblock \href {https://doi.org/10.18653/v1/N19-1423} {{BERT}: Pre-training of
  deep bidirectional transformers for language understanding}.
\newblock In \emph{Proceedings of the 2019 Conference of the North {A}merican
  Chapter of the Association for Computational Linguistics: Human Language
  Technologies, Volume 1 (Long and Short Papers)}, pages 4171--4186,
  Minneapolis, Minnesota. Association for Computational Linguistics.

\bibitem[{Dong and Sch{\"a}fer(2011)}]{dong2011ensemble}
Cailing Dong and Ulrich Sch{\"a}fer. 2011.
\newblock Ensemble-style self-training on citation classification.
\newblock In \emph{Proceedings of 5th international joint conference on natural
  language processing}, pages 623--631.

\bibitem[{Ge et~al.(2021)Ge, Dinh, Liu, Su, Lu, Wang, and
  Diesner}]{ge-etal-2021-baco}
Yubin Ge, Ly~Dinh, Xiaofeng Liu, Jinsong Su, Ziyao Lu, Ante Wang, and Jana
  Diesner. 2021.
\newblock \href {https://doi.org/10.18653/v1/2021.acl-long.116} {{BACO}: A
  background knowledge- and content-based framework for citing sentence
  generation}.
\newblock In \emph{Proceedings of the 59th Annual Meeting of the Association
  for Computational Linguistics and the 11th International Joint Conference on
  Natural Language Processing (Volume 1: Long Papers)}, pages 1466--1478,
  Online. Association for Computational Linguistics.

\bibitem[{Guo et~al.(2010)Guo, Korhonen, Liakata, Karolinska, Sun, and
  Stenius}]{guo2010identifying}
Yufan Guo, Anna Korhonen, Maria Liakata, Ilona~Silins Karolinska, Lin Sun, and
  Ulla Stenius. 2010.
\newblock Identifying the information structure of scientific abstracts: an
  investigation of three different schemes.
\newblock In \emph{Proceedings of the 2010 Workshop on Biomedical Natural
  Language Processing}, pages 99--107. Association for Computational
  Linguistics.

\bibitem[{Han et~al.(2001)Han, Pei, Mortazavi-Asl, Pinto, Chen, Dayal, and
  Hsu}]{han2001prefixspan}
Jiawei Han, Jian Pei, Behzad Mortazavi-Asl, Helen Pinto, Qiming Chen, Umeshwar
  Dayal, and Meichun Hsu. 2001.
\newblock Prefixspan: Mining sequential patterns efficiently by
  prefix-projected pattern growth.
\newblock In \emph{proceedings of the 17th international conference on data
  engineering}, pages 215--224. Citeseer.

\bibitem[{Hirohata et~al.(2008)Hirohata, Okazaki, Ananiadou, and
  Ishizuka}]{hirohata2008identifying}
Kenji Hirohata, Naoaki Okazaki, Sophia Ananiadou, and Mitsuru Ishizuka. 2008.
\newblock Identifying sections in scientific abstracts using conditional random
  fields.
\newblock In \emph{Proceedings of the Third International Joint Conference on
  Natural Language Processing: Volume-I}.

\bibitem[{Hoang and Kan(2010)}]{hoang2010towards}
Cong Duy~Vu Hoang and Min-Yen Kan. 2010.
\newblock Towards automated related work summarization.
\newblock In \emph{Coling 2010: Posters}, pages 427--435.

\bibitem[{Hu and Wan(2014)}]{hu2014automatic}
Yue Hu and Xiaojun Wan. 2014.
\newblock Automatic generation of related work sections in scientific papers:
  an optimization approach.
\newblock In \emph{Proceedings of the 2014 Conference on Empirical Methods in
  Natural Language Processing (EMNLP)}, pages 1624--1633.

\bibitem[{Huang et~al.(2020)Huang, Huang, Ding, Hsu, and
  Giles}]{huang-etal-2020-coda}
Ting-Hao~Kenneth Huang, Chieh-Yang Huang, Chien-Kuang~Cornelia Ding, Yen-Chia
  Hsu, and C.~Lee Giles. 2020.
\newblock \href {https://aclanthology.org/2020.nlpcovid19-acl.6} {{CODA-19}:
  Using a non-expert crowd to annotate research aspects on 10,000+ abstracts in
  the {COVID-19} open research dataset}.
\newblock In \emph{Proceedings of the 1st Workshop on {NLP} for {COVID-19} at
  {ACL} 2020}, Online. Association for Computational Linguistics.

\bibitem[{Jaidka et~al.(2018)Jaidka, Chandrasekaran, Rustagi, and
  Kan}]{jaidka2018insights}
Kokil Jaidka, Muthu~Kumar Chandrasekaran, Sajal Rustagi, and Min-Yen Kan. 2018.
\newblock Insights from cl-scisumm 2016: the faceted scientific document
  summarization shared task.
\newblock \emph{International Journal on Digital Libraries}, 19(2):163--171.

\bibitem[{Jaidka et~al.(2010)Jaidka, Khoo, and Na}]{jaidka2010imitating}
Kokil Jaidka, Christopher Khoo, and Jin-Cheon Na. 2010.
\newblock Imitating human literature review writing: An approach to
  multi-document summarization.
\newblock In \emph{International Conference on Asian Digital Libraries}, pages
  116--119. Springer.

\bibitem[{Jaidka et~al.(2013)Jaidka, Khoo, and Na}]{jaidka2013literature}
Kokil Jaidka, Christopher~SG Khoo, and Jin-Cheon Na. 2013.
\newblock Literature review writing: how information is selected and
  transformed.
\newblock In \emph{Aslib Proceedings}. Emerald Group Publishing Limited.

\bibitem[{Jaidka et~al.(2019)Jaidka, Yasunaga, Chandrasekaran, Radev, and
  Kan}]{jaidka2019cl}
Kokil Jaidka, Michihiro Yasunaga, Muthu~Kumar Chandrasekaran, Dragomir Radev,
  and Min-Yen Kan. 2019.
\newblock The cl-scisumm shared task 2018: Results and key insights.
\newblock \emph{arXiv preprint arXiv:1909.00764}.

\bibitem[{Jaidka et~al.(2011)Jaidka, Khoo, and Na}]{jaidka2011literature}
Kokil~Jaidka Jaidka, Christopher~Khoo Khoo, and Jin-Cheon~Na Na. 2011.
\newblock Literature review writing: a study of information selection from
  cited papers/kokil jaidka, christopher khoo and jin-cheon na.

\bibitem[{Jurgens et~al.(2018)Jurgens, Kumar, Hoover, McFarland, and
  Jurafsky}]{jurgens2018measuring}
David Jurgens, Srijan Kumar, Raine Hoover, Dan McFarland, and Dan Jurafsky.
  2018.
\newblock Measuring the evolution of a scientific field through citation
  frames.
\newblock \emph{Transactions of the Association for Computational Linguistics},
  6:391--406.

\bibitem[{Kan et~al.(2001)Kan, McKeown, and Klavans}]{kan-etal-2001-applying}
Min-Yen Kan, Kathleen~R. McKeown, and Judith~L. Klavans. 2001.
\newblock \href {https://aclanthology.org/W01-0813} {Applying natural language
  generation to indicative summarization}.
\newblock In \emph{Proceedings of the {ACL} 2001 Eighth {E}uropean Workshop on
  Natural Language Generation ({EWNLG})}, Toulouse, France. Association for
  Computational Linguistics.

\bibitem[{Khoo et~al.(2011)Khoo, Na, and Jaidka}]{khoo2011analysis}
Christopher~SG Khoo, Jin-Cheon Na, and Kokil Jaidka. 2011.
\newblock Analysis of the macro-level discourse structure of literature
  reviews.
\newblock \emph{Online Information Review}.

\bibitem[{Lauscher et~al.(2021)Lauscher, Ko, Kuhl, Johnson, Jurgens, Cohan, and
  Lo}]{lauscher2021multicite}
Anne Lauscher, Brandon Ko, Bailey Kuhl, Sophie Johnson, David Jurgens, Arman
  Cohan, and Kyle Lo. 2021.
\newblock Multicite: Modeling realistic citations requires moving beyond the
  single-sentence single-label setting.
\newblock \emph{arXiv preprint arXiv:2107.00414}.

\bibitem[{Lewis et~al.(2020)Lewis, Liu, Goyal, Ghazvininejad, Mohamed, Levy,
  Stoyanov, and Zettlemoyer}]{lewis-etal-2020-bart}
Mike Lewis, Yinhan Liu, Naman Goyal, Marjan Ghazvininejad, Abdelrahman Mohamed,
  Omer Levy, Veselin Stoyanov, and Luke Zettlemoyer. 2020.
\newblock \href {https://doi.org/10.18653/v1/2020.acl-main.703} {{BART}:
  Denoising sequence-to-sequence pre-training for natural language generation,
  translation, and comprehension}.
\newblock In \emph{Proceedings of the 58th Annual Meeting of the Association
  for Computational Linguistics}, pages 7871--7880, Online. Association for
  Computational Linguistics.

\bibitem[{Li and Wang(2008)}]{li2008efficiently}
Chun Li and Jianyong Wang. 2008.
\newblock Efficiently mining closed subsequences with gap constraints.
\newblock In \emph{proceedings of the 2008 SIAM International Conference on
  Data Mining}, pages 313--322. SIAM.

\bibitem[{Li et~al.(2020)Li, Burns, and Peng}]{li2020paragraph}
Xiangci Li, Gully Burns, and Nanyun Peng. 2020.
\newblock A paragraph-level multi-task learning model for scientific
  fact-verification.
\newblock \emph{arXiv preprint arXiv:2012.14500}.

\bibitem[{Li et~al.(2021)Li, Burns, and Peng}]{li2021scientific}
Xiangci Li, Gully Burns, and Nanyun Peng. 2021.
\newblock Scientific discourse tagging for evidence extraction.
\newblock In \emph{Proceedings of the 16th Conference of the European Chapter
  of the Association for Computational Linguistics: Main Volume}, pages
  2550--2562.

\bibitem[{Li and Ouyang(2022)}]{li2022automatic}
Xiangci Li and Jessica Ouyang. 2022.
\newblock Automatic related work generation: A meta study.
\newblock \emph{arXiv preprint arXiv:2201.01880}.

\bibitem[{Liakata(2010)}]{liakata2010zones}
Maria Liakata. 2010.
\newblock Zones of conceptualisation in scientific papers: a window to negative
  and speculative statements.
\newblock In \emph{Proceedings of the Workshop on Negation and Speculation in
  Natural Language Processing}, pages 1--4. Association for Computational
  Linguistics.

\bibitem[{Liakata et~al.(2012)Liakata, Saha, Dobnik, Batchelor, and
  Rebholz-Schuhmann}]{liakata2012automatic}
Maria Liakata, Shyamasree Saha, Simon Dobnik, Colin Batchelor, and Dietrich
  Rebholz-Schuhmann. 2012.
\newblock Automatic recognition of conceptualization zones in scientific
  articles and two life science applications.
\newblock \emph{Bioinformatics}, 28(7):991--1000.

\bibitem[{Lin(2004)}]{lin2004rouge}
Chin-Yew Lin. 2004.
\newblock Rouge: A package for automatic evaluation of summaries.
\newblock In \emph{Text summarization branches out}, pages 74--81.

\bibitem[{Liu et~al.(2019)Liu, Ott, Goyal, Du, Joshi, Chen, Levy, Lewis,
  Zettlemoyer, and Stoyanov}]{liu2019roberta}
Yinhan Liu, Myle Ott, Naman Goyal, Jingfei Du, Mandar Joshi, Danqi Chen, Omer
  Levy, Mike Lewis, Luke Zettlemoyer, and Veselin Stoyanov. 2019.
\newblock Roberta: A robustly optimized bert pretraining approach.
\newblock \emph{arXiv preprint arXiv:1907.11692}.

\bibitem[{Lo et~al.(2020)Lo, Wang, Neumann, Kinney, and
  Weld}]{lo-wang-2020-s2orc}
Kyle Lo, Lucy~Lu Wang, Mark Neumann, Rodney Kinney, and Daniel Weld. 2020.
\newblock \href {https://doi.org/10.18653/v1/2020.acl-main.447} {{S}2{ORC}: The
  semantic scholar open research corpus}.
\newblock In \emph{Proceedings of the 58th Annual Meeting of the Association
  for Computational Linguistics}, pages 4969--4983, Online. Association for
  Computational Linguistics.

\bibitem[{Luu et~al.(2021)Luu, Wu, Koncel-Kedziorski, Lo, Cachola, and
  Smith}]{luu-etal-2021-explaining}
Kelvin Luu, Xinyi Wu, Rik Koncel-Kedziorski, Kyle Lo, Isabel Cachola, and
  Noah~A. Smith. 2021.
\newblock \href {https://doi.org/10.18653/v1/2021.acl-long.166} {Explaining
  relationships between scientific documents}.
\newblock In \emph{Proceedings of the 59th Annual Meeting of the Association
  for Computational Linguistics and the 11th International Joint Conference on
  Natural Language Processing (Volume 1: Long Papers)}, pages 2130--2144,
  Online. Association for Computational Linguistics.

\bibitem[{Radford et~al.(2019)Radford, Wu, Child, Luan, Amodei, Sutskever
  et~al.}]{radford2019language}
Alec Radford, Jeffrey Wu, Rewon Child, David Luan, Dario Amodei, Ilya
  Sutskever, et~al. 2019.
\newblock Language models are unsupervised multitask learners.
\newblock \emph{OpenAI blog}, 1(8):9.

\bibitem[{Raffel et~al.(2020)Raffel, Shazeer, Roberts, Lee, Narang, Matena,
  Zhou, Li, and Liu}]{2020t5}
Colin Raffel, Noam Shazeer, Adam Roberts, Katherine Lee, Sharan Narang, Michael
  Matena, Yanqi Zhou, Wei Li, and Peter~J. Liu. 2020.
\newblock \href {http://jmlr.org/papers/v21/20-074.html} {Exploring the limits
  of transfer learning with a unified text-to-text transformer}.
\newblock \emph{Journal of Machine Learning Research}, 21(140):1--67.

\bibitem[{Ravi et~al.(2018)Ravi, Setlur, Ravi, and
  Govindaraju}]{ravi2018article}
Kumar Ravi, Srirangaraj Setlur, Vadlamani Ravi, and Venu Govindaraju. 2018.
\newblock Article citation sentiment analysis using deep learning.
\newblock In \emph{2018 IEEE 17th International Conference on Cognitive
  Informatics \& Cognitive Computing (ICCI* CC)}, pages 78--85. IEEE.

\bibitem[{Sang and Veenstra(1999)}]{sang1999representing}
Erik~F Sang and Jorn Veenstra. 1999.
\newblock Representing text chunks.
\newblock In \emph{Proceedings of the ninth conference on European chapter of
  the Association for Computational Linguistics}, pages 173--179. Association
  for Computational Linguistics.

\bibitem[{See et~al.(2017)See, Liu, and Manning}]{see-etal-2017-get}
Abigail See, Peter~J. Liu, and Christopher~D. Manning. 2017.
\newblock \href {https://doi.org/10.18653/v1/P17-1099} {Get to the point:
  Summarization with pointer-generator networks}.
\newblock In \emph{Proceedings of the 55th Annual Meeting of the Association
  for Computational Linguistics (Volume 1: Long Papers)}, pages 1073--1083,
  Vancouver, Canada. Association for Computational Linguistics.

\bibitem[{Stenetorp et~al.(2012)Stenetorp, Pyysalo, Topi{\'c}, Ohta, Ananiadou,
  and Tsujii}]{stenetorp2012brat}
Pontus Stenetorp, Sampo Pyysalo, Goran Topi{\'c}, Tomoko Ohta, Sophia
  Ananiadou, and Jun’ichi Tsujii. 2012.
\newblock Brat: a web-based tool for nlp-assisted text annotation.
\newblock In \emph{Proceedings of the Demonstrations at the 13th Conference of
  the European Chapter of the Association for Computational Linguistics}, pages
  102--107.

\bibitem[{Teufel and Moens(1999)}]{teufel1999discourse}
Simone Teufel and Marc Moens. 1999.
\newblock Discourse-level argumentation in scientific articles: human and
  automatic annotation.
\newblock \emph{Towards Standards and Tools for Discourse Tagging}.

\bibitem[{Teufel and Moens(2002)}]{teufel2002summarizing}
Simone Teufel and Marc Moens. 2002.
\newblock Summarizing scientific articles: experiments with relevance and
  rhetorical status.
\newblock \emph{Computational linguistics}, 28(4):409--445.

\bibitem[{Teufel et~al.(2006)Teufel, Siddharthan, and
  Tidhar}]{teufel2006automatic}
Simone Teufel, Advaith Siddharthan, and Dan Tidhar. 2006.
\newblock Automatic classification of citation function.
\newblock In \emph{Proceedings of the 2006 conference on empirical methods in
  natural language processing}, pages 103--110.

\bibitem[{Tuarob et~al.(2019)Tuarob, Kang, Wettayakorn, Pornprasit, Sachati,
  Hassan, and Haddawy}]{tuarob2019automatic}
Suppawong Tuarob, Sung~Woo Kang, Poom Wettayakorn, Chanatip Pornprasit,
  Tanakitti Sachati, Saeed-Ul Hassan, and Peter Haddawy. 2019.
\newblock Automatic classification of algorithm citation functions in
  scientific literature.
\newblock \emph{IEEE Transactions on Knowledge and Data Engineering},
  32(10):1881--1896.

\bibitem[{Vaswani et~al.(2017)Vaswani, Shazeer, Parmar, Uszkoreit, Jones,
  Gomez, Kaiser, and Polosukhin}]{vaswani2017attention}
Ashish Vaswani, Noam Shazeer, Niki Parmar, Jakob Uszkoreit, Llion Jones,
  Aidan~N Gomez, {\L}ukasz Kaiser, and Illia Polosukhin. 2017.
\newblock Attention is all you need.
\newblock In \emph{Advances in neural information processing systems}, pages
  5998--6008.

\bibitem[{Vyas et~al.(2020)Vyas, Ravi, Ravi, Uma, Setlur, and
  Govindaraju}]{vyas2020article}
Vishal Vyas, Kumar Ravi, Vadlamani Ravi, V~Uma, Srirangaraj Setlur, and Venu
  Govindaraju. 2020.
\newblock Article citation study: Context enhanced citation sentiment
  detection.
\newblock \emph{arXiv preprint arXiv:2005.04534}.

\bibitem[{Wang et~al.(2019)Wang, Li, Zhou, Tang, and Wang}]{wang2019toc}
Pancheng Wang, Shasha Li, Haifang Zhou, Jintao Tang, and Ting Wang. 2019.
\newblock Toc-rwg: Explore the combination of topic model and citation
  information for automatic related work generation.
\newblock \emph{IEEE Access}, 8:13043--13055.

\bibitem[{Wolf et~al.(2020)Wolf, Debut, Sanh, Chaumond, Delangue, Moi, Cistac,
  Rault, Louf, Funtowicz, Davison, Shleifer, von Platen, Ma, Jernite, Plu, Xu,
  Le~Scao, Gugger, Drame, Lhoest, and Rush}]{wolf-etal-2020-transformers}
Thomas Wolf, Lysandre Debut, Victor Sanh, Julien Chaumond, Clement Delangue,
  Anthony Moi, Pierric Cistac, Tim Rault, Remi Louf, Morgan Funtowicz, Joe
  Davison, Sam Shleifer, Patrick von Platen, Clara Ma, Yacine Jernite, Julien
  Plu, Canwen Xu, Teven Le~Scao, Sylvain Gugger, Mariama Drame, Quentin Lhoest,
  and Alexander Rush. 2020.
\newblock \href {https://doi.org/10.18653/v1/2020.emnlp-demos.6} {Transformers:
  State-of-the-art natural language processing}.
\newblock In \emph{Proceedings of the 2020 Conference on Empirical Methods in
  Natural Language Processing: System Demonstrations}, pages 38--45, Online.
  Association for Computational Linguistics.

\bibitem[{Xing et~al.(2020)Xing, Fan, and Wan}]{xing2020automatic}
Xinyu Xing, Xiaosheng Fan, and Xiaojun Wan. 2020.
\newblock Automatic generation of citation texts in scholarly papers: A pilot
  study.
\newblock In \emph{Proceedings of the 58th Annual Meeting of the Association
  for Computational Linguistics}, pages 6181--6190.

\bibitem[{Yasunaga et~al.(2019)Yasunaga, Kasai, Zhang, Fabbri, Li, Friedman,
  and Radev}]{yasunaga2019scisummnet}
Michihiro Yasunaga, Jungo Kasai, Rui Zhang, Alexander~R Fabbri, Irene Li, Dan
  Friedman, and Dragomir~R Radev. 2019.
\newblock Scisummnet: A large annotated corpus and content-impact models for
  scientific paper summarization with citation networks.
\newblock In \emph{Proceedings of the AAAI Conference on Artificial
  Intelligence}, volume~33, pages 7386--7393.

\bibitem[{Yasunaga et~al.(2017)Yasunaga, Zhang, Meelu, Pareek, Srinivasan, and
  Radev}]{yasunaga2017graph}
Michihiro Yasunaga, Rui Zhang, Kshitijh Meelu, Ayush Pareek, Krishnan
  Srinivasan, and Dragomir Radev. 2017.
\newblock Graph-based neural multi-document summarization.
\newblock In \emph{Proceedings of the 21st Conference on Computational Natural
  Language Learning (CoNLL 2017)}, pages 452--462.

\end{thebibliography}
\bibliographystyle{acl_natbib}
\clearpage
\appendix

\section{Appendix}
\label{sec:appendix}

\begin{table*}[t]
\begin{center}
\small
    \begin{tabular}{  l | l  l  l   | l  l  l }
    \hline
     & \multicolumn{3}{c}{\textit{Five-fold cross-validation scores}} & \multicolumn{3}{c}{\textit{Test-set scores}} \\
    \textbf{Models} & \textbf{Disc} & \textbf{CT}  & \textbf{CS} & \textbf{Disc} & \textbf{CT} & \textbf{CS} \\ \hline
    SciBERT \cite{beltagy-etal-2019-scibert} & $\textbf{0.900}$ ($0.0099$) & $\textbf{0.961}$ ($0.0038$) & $\textbf{0.926}$ ($0.0059$) & $\textbf{0.898}$ & $\textbf{0.959}$ & $\textbf{0.930}$\\ \hline

    Roberta-base \cite{liu2019roberta} & $0.886$  ($0.0050$) & $0.956$  ($0.0036$) & $0.922$  ($0.0048$) & $0.885$ & $0.956$ & $0.929$  \\ \hline
    
    BERT-base \cite{devlin-etal-2019-bert} & $0.879$ ($0.0070$) & $0.954$  ($0.0055$) & $0.910$  ($0.0064$) & $0.875$ &  $0.952$ & $0.915$  \\ \hline
    
    LED-base (Pretrained) & $0.872$ ($0.0253$)  & $0.948$ ($0.0117$)  & $0.905$  ($0.0088$) & $0.869$ & $0.910$ & $0.907$  \\ 

    LED-base \cite{Beltagy2020Longformer}  & $0.865$  ($0.0090$) & $0.922$  ($0.0128$) & $0.907$  ($0.0074$) & $0.842$ & $0.874$ & $0.909$  \\ \hline

    \end{tabular}
    \vspace{-0.5em}
    \caption{Micro-F1 scores for the joint related work tagger using different language models as the encoder. The tasks are discourse tagging (Disc), citation type recognition (CT), and citation span detection (CS). Five-fold cross-validation scores are reported as the mean (standard deviation) across all folds. The pretraining of LED is explained in \cref{sec:span_geneartion_setting}.} \label{tab:full_joint_tagger_performance}
    \vspace{-1.5em}
\end{center}
\end{table*}

\subsection{Training Configurations}
For the joint related work tagger training, we use GeForce GTX 1080 11 GB GPUs. The training process lasts 2.5 hours on a single GPU using Huggingface's \cite{wolf-etal-2020-transformers} SciBERT, BERT-base or Roberta-base as the paragraph encoders, and it lasts 6.5 hours using LED-base encoder. We train the models for 15 epochs. It takes approximately one week to run the hyper-parameter search using five-fold cross-validation for all language models, using 8 GPUs in total. 

For training the citation span generation model, we use Tesla V100s-PCIE-32GB GPUs. The training process lasts for 2 days on a single GPU. We run the training for a maximum of 3 epochs with early stopping based on the validation loss.

\subsection{Ethical Considerations}
We present a new dataset that is derived from the S2ORC dataset \cite{lo-wang-2020-s2orc}, which is released under CC BY-NC 2.0 license. The Huggingface models \cite{wolf-etal-2020-transformers} we develop upon are released under Apache License 2.0.

Our annotators were compensated for their work at a rate of double the minimum wage in our local area.

\begin{table*}[t]
\begin{center}
\small
    \begin{tabular}{ l }
    \hline
    \textbf{Discourse Subsequence} \\
    \emph{transition}, \emph{narrative\_cite}, \emph{single\_summ} \\
    \textbf{Functionalities} \\ 
    Introducing an approach and providing background knowledge. \\
    \textbf{Examples} \\
    1. Joint POS tagging with parsing is not a new idea. \\ 
    2. In PCFG-based parsing (Collins, 1999; Charniak, 2000; Petrov et al., 2006), POS tagging is \\ 
       considered as a natural step of parsing by employing lexical rules. \\
    3. For transition-based parsing, Hatori et al. (2011) proposed to integrate POS tagging with \\ 
        dependency parsing. \\ 
    \hline \hline
    \textbf{Discourse Subsequence} \\
    \emph{single\_summ}, \emph{reflection} \\
    \textbf{Functionalities} \\ 
    Comparing the prior work to the current work. \\
    \textbf{Examples} \\
    1. Haghighi et al. (2009) confirm and extend these results, showing BLEU improvement for \\ 
      a hierarchical phrase-based MT system on a small Chinese corpus. \\ 
    2. As opposed to ITG, we use a linguistically motivated phrase-structure tree to drive our search\\ 
     and inform our model. \\
    \hline \hline
    \textbf{Discourse Subsequence} \\
    \emph{reflection}, \emph{single\_summ}\\
    \textbf{Functionalities} \\ 
    Supporting the current work with a previous work. \\
    \textbf{Examples} \\
    1. Our baseline semi-supervised model can be viewed as an extension of these approaches to a \\ 
    reading comprehension setting. \\ 
    2.	Dai et al. (2015) also explore initialization from a language model, but find that the \\
    recurrent autoencoder is superior, which is why we do not consider language models in this work. \\
    \hline \hline
    \textbf{Discourse Subsequence} \\
    \emph{transition}, \emph{narrative\_cite}, \emph{transition}\\
    \textbf{Functionalities} \\ 
    Topic sentence, narration of prior work followed by critique. \\
    \textbf{Examples} \\
    1.	Traditional work on relation classification can be categorized into feature-based methods \\
    and kernel-based methods. \\ 
    2. The former relies on a large number of human-designed features (Zhou et al., 2005; Jiang and \\ 
    Zhai, 2007; Li and Ji, 2014) while the latter leverages various kernels to implicitly explore a much \\ 
    larger feature space (Bunescu and Mooney, 2005; Nguyen et al., 2009 ). \\
    3.	However, both methods suffer from error propagation problems and poor generalization abilities \\
    on unseen words. \\
    \hline 
    
    \end{tabular}
    \vspace{-0.5em}
    \caption{Frequent discourse label subsequences detected by applying PrefixSpan \cite{han2001prefixspan} and Gap-Bide algorithm \cite{li2008efficiently}.} \label{tab:discourse_subsequence}
    \vspace{-1.5em}
\end{center}
\end{table*}

\begin{table*}[t]
\begin{center}
\small
    \begin{tabular}{ l }
    \hline
    \textbf{Discourse Subsequence} \\
    \emph{single\_summ}, \emph{single\_summ}, \emph{transition}\\
    \textbf{Functionalities} \\ 
    Commenting previous works summarized. \\
    \textbf{Examples} \\
    1. Walker et al. (2012) extract rules representing characters from their annotated movie \\ 
    subtitle corpora. \\ 
    2. Miyazaki et al. (2015) propose a method of converting utterances using rewriting rules \\ 
    automatically derived from a Twitter corpus. \\
    3. These approaches have a fundamental problem to need some manual annotations, which is a \\ 
    main issue to be solved in this work. \\
    \hline \hline 
    \textbf{Discourse Subsequence} \\
    \emph{narrative\_cite}, \emph{transition}, \emph{single\_summ}\\
    \textbf{Functionalities} \\ 
    Criticizing the previously cited work and citing an improved work. \\
    \textbf{Examples} \\
    1. There have also been several classical studies based on nonneural approaches to headline \\ 
    generation (Woodsend et al., 2010; Alfonseca et al., 2013; Colmenares et al., 2015) , \\
    but they basically addressed sentence compression after extracting important linguistic \\ 
    units such as phrases. \\
    2. In other words, their methods can still yield erroneous output, although they would be more \\ 
     controllable than neural models. \\
    3. One exception is the work of Alotaiby (2011) , where fixed-sized substrings were considered \\
    for headline generation. \\
    \hline
    \textbf{Discourse Subsequence} \\
    \emph{narrative\_cite}, \emph{transition}, \emph{single\_summ}\\
    \textbf{Functionalities} \\ 
    Describing an idea following by a comment and then citations implementing the idea. \\
    \textbf{Examples} \\
    1. One of the classes of errors in the Helping Our Own (HOO) 2011 shared task (Dale and  \\ 
    Kilgarriff, 2011) was punctuation.\\
    2. Comma errors are the most frequent kind of punctuation error made by learners. \\
    3. Israel et al. (2012) present a model for detecting these kinds of errors in learner texts. \\
    \hline
    \end{tabular}
    \vspace{-0.5em}
    \caption{Frequent discourse label subsequences detected by applying PrefixSpan \cite{han2001prefixspan} and Gap-Bide algorithm \cite{li2008efficiently}, continued.} \label{tab:discourse_subsequence_countinue}
    \vspace{-1.5em}
\end{center}
\end{table*}

\end{document}